\documentclass{article}

\usepackage{arxiv}

\usepackage[utf8]{inputenc} 
\usepackage[T1]{fontenc}    
\usepackage{hyperref}       
\usepackage{url}            
\usepackage{booktabs}       
\usepackage{amsfonts}       
\usepackage{nicefrac}       
\usepackage{microtype}      
\usepackage{lipsum}		
\usepackage{graphicx}
\usepackage{natbib}
\usepackage{doi}

\usepackage{graphicx}%
\usepackage{multirow}%
\usepackage{amsmath,amssymb,amsfonts}%
\usepackage{amsthm}%
\usepackage{mathrsfs}%
\usepackage[title]{appendix}%
\usepackage{xcolor}%
\usepackage{textcomp}%
\usepackage{manyfoot}%
\usepackage{booktabs}%
\usepackage{algorithm}%
\usepackage{algorithmicx}%
\usepackage{algpseudocode}%
\usepackage{listings}%
\usepackage{tikz}
\usepackage{pgfplots}

\title{FarsiMCQGen: a Persian Multiple-choice Question Generation Framework}

\author{
	Mohammad Heydari Rad \\
	Department of Computer Engineering\\
	Amirkabir University of Technology\\
	\texttt{mhrad81@aut.ac.ir} \\
	\And
	Rezvan Afari \\
	Department of Computer Engineering\\
	Amirkabir University of Technology\\
	\texttt{ghba@aut.ac.ir} \\
	\And
	Saeedeh Momtazi \\
	Department of Computer Engineering\\
	Amirkabir University of Technology\\
	\texttt{momtazi@aut.ac.ir} \\
}

\date{}

\renewcommand{\headeright}{}
\renewcommand{\undertitle}{}
\renewcommand{\shorttitle}{}

\hypersetup{
pdftitle={FarsiMCQGen: a Persian Multiple-choice Question Generation Framework},
pdfsubject={Multiple-choice Question Generation},
pdfauthor={Mohammad Heydari Rad, Rezvan Afari, Saeedeh Momtazi},
pdfkeywords={BERT, Fill-Mask, Knowledge Graph, Persian Dataset, Multiple-choice},
}

\begin{document}
\maketitle

\begin{abstract}
Multiple-choice questions (MCQs) are commonly used in educational testing, as they offer an efficient means of evaluating learners’ knowledge. However, generating high-quality MCQs, particularly in low-resource languages such as Persian, remains a significant challenge. This paper introduces \textbf{FarsiMCQGen}, an innovative approach for generating Persian-language MCQs. Our methodology combines candidate generation, filtering, and ranking techniques to build a model that generates answer choices resembling those in real MCQs. We leverage advanced methods, including Transformers and knowledge graphs, integrated with rule-based approaches to craft credible distractors that challenge test-takers. Our work is based on data from Wikipedia, which includes general knowledge questions. Furthermore, this study introduces a novel Persian MCQ dataset comprising 10,289 questions. This dataset is evaluated by different state-of-the-art large language models (LLMs).
Our results demonstrate the effectiveness of our model and the quality of the generated dataset, which has the potential to inspire further research on MCQs.
\end{abstract}


\section{Introduction}
\label{sec:introduction}
Multiple-choice Questions (MCQs) are widely used in educational assessments, valued for allowing fast and automatic evaluation of students' understanding and their applicability to large-scale tests, such as university entrance exams. However, manually creating MCQs is often time-consuming and requires significant expertise. To address these limitations, researchers have turned to artificial intelligence (AI) for automatic question generation. Early automatic question generation systems relied primarily on rule-based methods, but recent advances in deep learning have led to the emergence of Neural Question Generation (NQG). NQG leverages neural network-based models to automatically generate questions from textual input, offering a more scalable and accurate solution \citep{faraby2023nqg, guo2024nqgsurvey}.

Advances in Natural Language Processing (NLP) have revolutionized question and answer related applications, particularly in education. NLP techniques now support a wide range of tasks, including generating questions from text \citep{deepak2019ontoquest}, reading comprehension \citep{rogers2023qa, riza2023automatic}, cloze tests \citep{qiu2021automatic,hoshino2008cloze, matsumori2023mask, kumar2024novelmcq, johnshon2024aqgspanish}, Wh-type questions \citep{kumar2024novelmcq}, and generating distractors \citep{gao2019generating, baldwin2022natural}. By automating such processes that traditionally rely on human effort, NLP has the potential to reduce costs and enhance educational practices.

Many researchers have made progress in Persian question answering by fine-tuning different Transformer-based models like 
\texttt{xlm-roberta-large-fa-qa} \footnote{\href{https://huggingface.co/SajjadAyoubi/xlm-roberta-large-fa-qa}{Hugging Face: xlm-roberta-large-fa-qa}} and \texttt{bert-base-fa-qa} \footnote{\href{https://huggingface.co/SajjadAyoubi/bert-base-fa-qa}{Hugging Face: bert-base-fa-qa}} \citep{PersianQA} for question answering, and \texttt{mT5-large-parsinlu-multiple-choice} \footnote{\href{https://huggingface.co/persiannlp/mt5-large-parsinlu-multiple-choice}{Hugging Face: mT5-large-parsinlu-multiple-choice}} and \texttt{mt5-base-parsinlu-arc-comqa-obqa-multiple-choice} \footnote{\href{https://huggingface.co/persiannlp/mt5-base-parsinlu-arc-comqa-obqa-multiple-choice}{Hugging Face: mt5-base-parsinlu-arc-comqa-obqa-multiple-choice}} \citep{2020parsiglue} for answering MCQs. Moreover, neural question generation methods can be used for data augmentation to enhance QA systems \citep{faraby2023nqg}.

Furthermore, different question answering datasets have been created for the Persian language, including PQuAD \citep{darvishi2023pquad}, ParSQuAD \citep{Abadani}, PersianQuAD \citep{9729745}, PersianQA \citep{PersianQA}, PCoQA \citep{hemati2023pcoqa}, and PerCQA \citep{jamali-etal-2022-percqa}. Also, \citet{farsi2025persian} introduced a Persian Visual Question Answering Dataset. In the realm of Persian MCQ datasets, \citet{2020parsiglue}, \citet{farsi2025melac} and \citet{ghahroodi2024khayyamchallenge} mainly utilized existing sources (e.g. university entrance exam questions), and an automated framework for generating MCQs is still missing.

In this research, we designed an MCQ generation system consisting of two parts. The first part receives a text passage along with a short correct answer and generates a question based on the given information; the second part then receives a question with its correct answer and generates three wrong choices. Creating these wrong choices involves two important considerations: they need to be similar to the correct answer, and they must also make sense within the context of the question. Our approach for generating wrong choices is divided into three main stages: (1) generating a pool of potential candidates, (2) refining this pool by filtering out unsuitable candidates, and (3) ranking the remaining candidates to select the three best options as the final outputs.

The main contributions of this work are summarized as follows: (1) Proposing a lightweight framework for generating MCQs using language models and a knowledge graph, (2) Constructing a dataset through the proposed framework, which can serve as a benchmark for general knowledge assessment, and (3) Categorizing the generated questions based on their type and content, and evaluating them using state-of-the-art large language models (LLMs) and human evaluation.

The remainder of this paper is organized as follows. Section \ref{sec:relatedWorks} reviews related work in this field. Section \ref{sec:ProposedMethod} presents the proposed method for generating MCQs. Section \ref{sec:statistics} provides an overview of the dataset, including the types and content categories of the questions. Section \ref{sec:eval} evaluates the quality of the generated questions. Finally, Section \ref{sec:conclusion} concludes the paper.

\section{Related Work}
\label{sec:relatedWorks}
MCQ generation aims at creating wrong answer choices to challenge and engage test takers. MCQs are widely utilized in various applications, including cloze tests, reading comprehension assessments, and general knowledge quizzes. In this section, we explore prior research and notable contributions in the field of MCQ generation within the context of NLP. 

The initial works in this area often employed basic approaches. These approaches focused on generating wrong answer choices that exhibited similarity in spelling \citep{fremer1969computer} or meaning \citep{richards1966can}.
With the advent of new technologies and advanced databases, the field of MCQ generation has undergone significant developments. Due to the nature of MCQs, where the wrong answers should be similar to the correct answer, the first attempts to generate multiple choices were based on the WordNet database \citep{miller1990introduction}, WordNet is a lexical database that links words through semantic relations. \citet{mitkov2003computer} employed the WordNet database and used various NLP techniques such as term extraction and shallow parsing \citep{van2002shallow} to generate the MCQs. After that \citet{mitkov2009semantic} introduced a measure that considered semantic similarity, distributional similarity, and phonetic similarity based on WordNet, resulting in significant improvements in the quality of the generated distractors (wrong answers). 
The proposed models by \citet{lin2007automatic, susanti2015automatic, pino2008selection} are other examples of MCQ generation based on WordNet. 

Recently, neural network-based models have emerged in the field of NLP, specifically in MCQ generation. \citet{gao2019generating} developed a hierarchical encoder-decoder network. \citet{zhou2020co} further improved the approach by enhancing the interaction between the generated question and the passage. Additionally, \citet{qiu-etal-2020-automatic} employed a sequence-to-sequence model and incorporated attention layers. They introduced two reforming modules to highlight distractor-relevant words and constrain answer-relevant words by measuring the distances between them and the correct answer.

Recent research in this area widely utilized Transformers and pre-trained models. \citet{chung2020bert} employed the BERT model \citep{devlin-etal-2019-bert} and utilized a negative answer strategy to effectively generate multiple distractors. \citet{chiang2022cdgp} introduced a two-stage approach. In the first stage, the candidate distractors were generated using a pre-trained model. Then, a Distractor Selector component was employed to select the three best distractors. This framework extends the work by \citet{ren2021knowledge}, which generates candidates based on the concepts related to the answer using a probabilistic topic model. \citet{rodriguez2022end} generated the distractors using the T5-small model \citep{raffel2020exploring} which is trained on a DG-Race dataset \citep{gao2019generating}. \citet{10577164} introduced an approach using integration of LLMs with retrieval-augmented generation and prompt engineering to employ chain-of-thought with self-refinement to make generated questions more challenging. \citet{luo-etal-2024-chain} proposed a framework to generate MCQs from multimodal input, utilizing a multimodal language model to generate the question, rationale and distractors. \citet{yu-etal-2024-automating-true} employed the combination of pre-trained language models and sentence retrieval techniques to generate distractors for True-False MCQs.

In Persian, \citet{zeinalipour-etal-2025-persianmcq} introduced an instruction-tuning dataset for fine-tuning open-source LLMs to generate MCQs from text and fine-tuned three small-scaled models on their dataset. While their approach relies on a single LLM to generate both questions and choices, our approach employs multiple types of language models for choice generation, various approaches for filtering, and a knowledge graph combined with a BERT-based model for ranking generated choices.

\section{Proposed Method}
\label{sec:ProposedMethod}

In this section, the proposed framework for generating MCQs is described. The pipeline takes as input a text and a short answer span within that text, and generates a question along with a set of answer choices, treating the given span as the correct choice.

To construct the required question–answer pairs, we utilized the PQuAD dataset \citep{darvishi2023pquad}, a large-scale Persian dataset containing over 80,000 question–answer pairs derived from Wikipedia. We focused on entries with short answers and used the corresponding paragraphs and answers to generate MCQs, where the given answer served as the correct choice.

\begin{figure*}[h]
	\centering
	\includegraphics[scale=0.6]{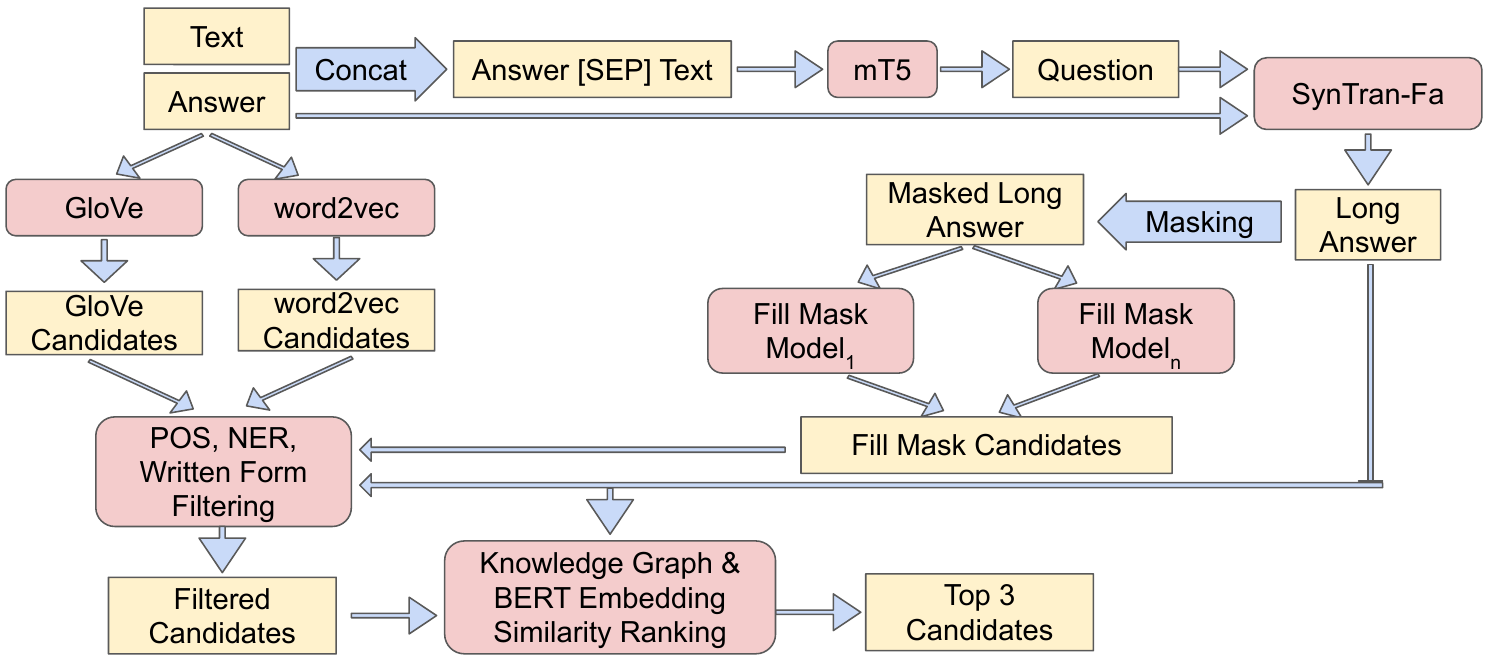}
	\caption{Question and wrong choices generation process}
	\label{fig:complete-process}
\end{figure*}

The overall architecture of the proposed framework is illustrated in Figure \ref{fig:complete-process}. The pipeline consists of two main components: (1) question generation, and (2) wrong choice generation.


\subsection{Generating Question from Input Text}


In this part, we aim at creating a question from the input text.
To develop an automatic model for generating questions from text, we employed the \texttt{mT5-base-finetuned-persian} model\footnote{\href{https://huggingface.co/eslamxm/mt5-base-finetuned-persian}{Hugging Face: mT5-base-finetuned-persian}}, a pretrained mT5 model \citep{xue-etal-2021-mt5} designed for multilingual generation tasks. We fine-tuned this model on the the PQuAD dataset \citep{darvishi2023pquad} by using the format answer + [SEP] + text as the input, where [SEP] is a special token used to separate the answer from the text. This setup allows the model to effectively generate the corresponding questions. By leveraging the strengths of the mT5 model, we established a foundation for generating contextually relevant questions.


\subsection{Generating Wrong Choices}
In the second part, we aim at generating wrong choices to be served with the target answer as multiple choices for the question. The wrong choice generation steps are as follows:

\begin{itemize}
	\item \textbf{Candidate Generation}: In the first step, two methods to generate candidate wrong choices are employed. First, we use the fill-mask technique with Transformer-based language models. Second, we identify the top 10 nearest words to the correct answer based on their GloVe \citep{pennington-etal-2014-glove} and Word2Vec \citep{mikolov2013efficient} embeddings, utilizing word embeddings from DadmaTools \citep{jafari-etal-2025-dadmatools}, which are trained on a Wikipedia corpus and well-aligned with our task.
	
	\item \textbf{Filtering}: In the second step, a filtering approach is used to eliminate unsuitable candidates. This approach helps the model to discard low-quality wrong choices, making the subsequent ranking step faster and more efficient.
	
	\item \textbf{Ranking and Selection}: Finally, we rank the filtered candidates and select the top three wrong choices. This ranking process ensures that the chosen wrong choices are the most plausible and contextually appropriate.
	
\end{itemize}

By following these three steps, we ensure the generation of high-quality MCQs, with wrong choices that are both semantically relevant and suitably challenging. All these steps are explained in detail in the following sections.

\subsubsection{Generating Wrong Choice Candidates}
In this step, we aim to generate wrong choices. Two key criteria are considered: the similarity of the wrong choices to the correct answer, and the relevance of the generated candidates with the question. This is crucial in cases where words with similar forms have different meanings; for example, "fly" can refer to both the act of flying as well as a type of insect. 
To address this challenge, we employed both fill-mask and similar embedding methods to generate the wrong candidates.

\begin{figure}[h]
	\centering
	\includegraphics[width=10cm, height=3.64cm]{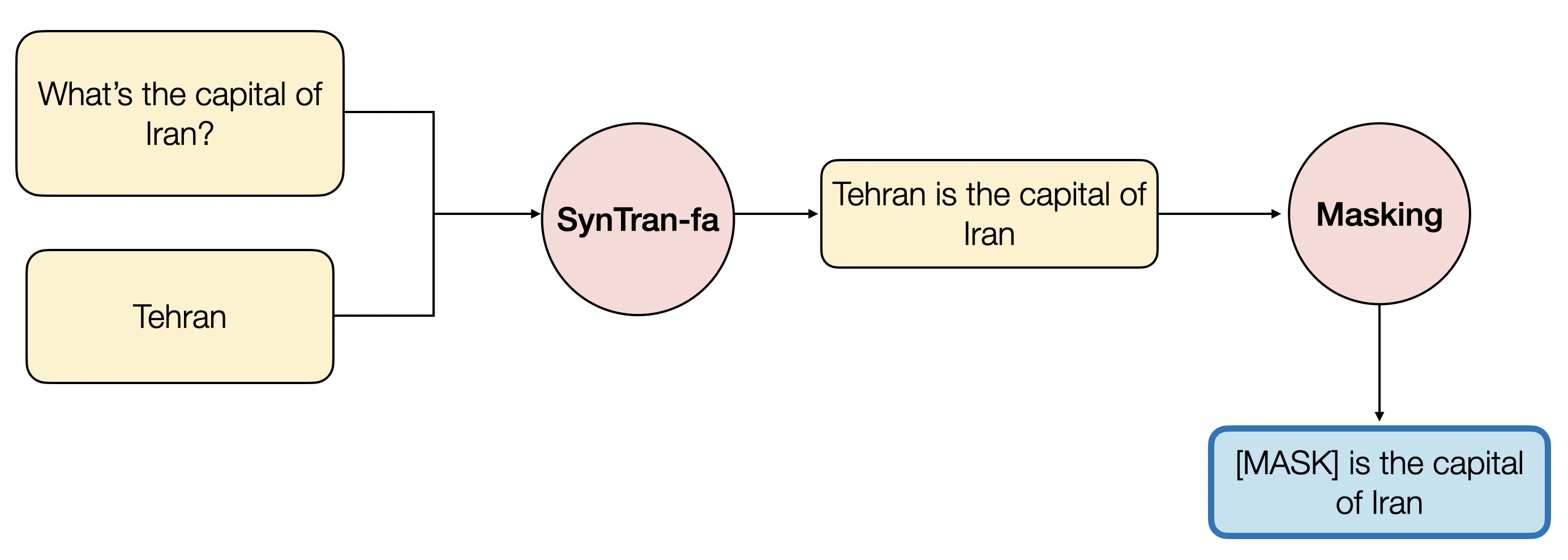}
	\caption{Generating answer sentence from question-answer pair and masking}
	\label{fig:fill-mask}
\end{figure}

\textbf{Fill-Mask Method:}
To effectively utilize the fill-mask method, a full sentence is required rather than a single word. To achieve this, we employed the SynTran-fa model \citep{202410.1684}, which constructs a complete answer sentence (long answer) from a given short answer and its corresponding question. By applying this model to our \texttt{(question, short answer)} pairs, we generate answer sentences. These long answers are also used for the steps that use context for filtering and ranking generated wrong choices. We then mask the short answer within these complete sentences and input the masked sentences into various models to predict the masked elements. For the fill-mask task, we leverage different Transformer-based models including ParsBERT\footnote{\href{https://huggingface.co/HooshvareLab/bert-base-parsbert-uncased}{Hugging Face: bert-base-parsbert-uncased}}, ParsBERT (v2.0)\footnote{\href{https://huggingface.co/HooshvareLab/bert-fa-base-uncased}{Hugging Face: bert-fa-base-uncased}}, ParsBERT (v3.0)\footnote{\href{https://huggingface.co/HooshvareLab/bert-fa-zwnj-base}{Hugging Face: bert-fa-zwnj-base}} \citep{Farahani2021}, ParsBigBird\footnote{\href{https://huggingface.co/SajjadAyoubi/distil-bigbird-fa-zwnj}{Hugging Face: distil-bigbird-fa-zwnj}} \citep{ParsBigBird}, ALBERT-Persian\footnote{\href{https://huggingface.co/m3hrdadfi/albert-fa-zwnj-base-v2}{Hugging Face: albert-fa-zwnj-base-v2}} \citep{ALBERTPersian},  ParsRoBERTa\footnote{\href{https://huggingface.co/HooshvareLab/roberta-fa-zwnj-base}{Hugging Face: roberta-fa-zwnj-base}} and DistilBERT\footnote{\href{https://huggingface.co/HooshvareLab/distilbert-fa-zwnj-base}{Hugging Face: distilbert-fa-zwnj-base}}. 
The process of generating answer sentences from question-answer pairs and masking is presented in Figure \ref{fig:fill-mask}.

Using this method ensures that the generated wrong choices are contextually relevant and syntactically fluent within the question framework. This dual consideration of context and fluency enhances the quality and fluency of the multiple-choice options, thereby improving the overall reliability and validity of the assessment. Additionally, by employing multiple state-of-the-art language models, we ensure a robust generation process capable of addressing the nuances of Persian language semantics. 

\textbf{Similar Embedding Method:}
To ensure that the generated candidates were similar to the correct answer, we employed a semantic word similarity approach. Alongside the wrong choices generated by the fill-mask method, we also identify the top 10 nearest words based on their GloVe and Word2Vec embeddings. We utilize word embeddings trained on a Wikipedia corpus, which perfectly suits our task as our questions are also based on the Wikipedia content. This combined approach guarantees that the candidate list included words that were semantically close to the correct answer, thereby enhancing the plausibility and quality of the wrong choices in the MCQs.

\subsubsection{Filtering}
To enhance the efficiency of the ranking step, a practical approach involves preliminary candidate filtering before presenting the candidate list for ranking. The filtering process includes the following components:

\textbf{POS Filter:} In certain cases, words can significantly impact the grammatical structure of sentences, potentially making the multiple-choice task more challenging. To generate better results, we streamline the candidate selection process by retaining only those candidates that share the same Part Of Speech (POS) tags as the correct answer in its position in the long answer. This approach can help ensure that the grammatical structure of the choices aligns with the correct answer, improving the quality of the multiple-choice task. To address this issue, we employed the Stanza library \citep{qi2020stanza}, a powerful tool capable of performing POS tagging on the provided sentences. To determine the appropriate choices for a given answer and question, both the universal POS and dependency relation are examined and taken into account.

To be more precise consider the following sentences: (1) "she went to school \underline{\textbf{yesterday}}", (2) "she went to school \underline{\textbf{fast}}". Both sentences are semantically and grammatically correct for "she went to school <mask>" but they are totally different together. Indeed, the pos-filter can be utilized to screen and filter specific types of choices, as demonstrated in the example above. This helps refine the selection process for more suitable options.

\textbf{Written Form Filter:} In some instances, representing numbers in word and digit is common; e.g., ``2'' and ``two'' , to resolve this issue and make the choices more uniform, we employed the Parsinorm library \citep{oji2021parsinorm}. This library helps standardize numerical representations by converting numbers between word and digit forms.

\textbf{NER Filter:} For all categories except \texttt{others} and \texttt{numbers} a Named Entity Recognition filter is applied using \texttt{bert-fa-zwnj-base-ner}\footnote{\href{https://huggingface.co/HooshvareLab/bert-fa-zwnj-base-ner}{Hugging Face: bert-fa-zwnj-base-ner}} \citep{ParsNER}. Candidates that do not share the same named entity type as the correct answer are discarded. Furthermore, for answers falling within the \texttt{numbers} category, only those that solely contained digits were selected.

\subsubsection{Ranking}

To rank generated candidates, two different similarity approaches are employed which are explained in this section.

\textbf{Knowledge Graph Embedding Similarity:}
In this module, we utilize a knowledge graph to enhance the generation and ranking of wrong choices. 
To this aim, we use FarsWikiKG \citep{shirmardi2021farswikikg}, a knowledge graph extracted from Wikipedia infoboxes.
Having a knowledge graph, we focused on the corresponding nodes to the candidate entities, including the correct choice as well as the wrong choices, and use their embeddings by applying Complex method \citep{10.5555/3045390.3045609} to find similarity\footnote{Cosine similarity is used to calculate these similarities.} between these target nodes. High similarities between nodes signify that these entities share common relationships or attributes. By integrating this module into our model, we significantly augment its understanding and reasoning capabilities, since it provides a vast contextual background, enabling our model to make more informed decisions.

\textbf{BERT Similarity:}
To calculate the semantic similarity between each wrong choice and the correct answer in the given context, we employed ParsBERT \citep{Farahani2021}, as a pre-trained BERT model for Persian. To this aim, we extract the word embeddings from the second layer output. Subsequently, the cosine similarity between the word embeddings of the correct answer and each wrong choice is calculated. By utilizing this module, our model is able to access more comprehensive information about the question. This approach is particularly valuable due to BERT's contextual word embedding method, which takes into account the context rather than individual words.

Afterward, we utilize two key features: the BERT similarity score and the similarity score derived from the knowledge graph. First, we normalize the scores of each feature to ensure comparability. The filtered results are then ranked based on the average of these two normalized scores. This approach allows us to combine the semantic similarity with the rich, contextual relationships captured in the knowledge graph. Finally, the top three most relevant items for each given question and answer were selected as the wrong choices, ensuring that they were both contextually appropriate and semantically plausible.

\section{Statistics}
\label{sec:statistics}
This section analyzes the dataset based on question type and content to highlight its diversity and coverage across different knowledge domains.

\subsection{Categorizing Based on Type}

\begin{figure}[h]
	\centering
	\begin{tikzpicture}
		\begin{axis}[
			ybar,
			symbolic x coords={What, When, How, How many, Where, Who, Which},
			ymin=0, ymax=3200,
			xtick=data,
			nodes near coords,
			bar width=0.5cm,
			width=11cm, height=3.7cm,
			xticklabel style={rotate=25, anchor=east},
			ylabel={Frequency},
			ylabel style={yshift=0.2cm},
			xlabel style={yshift=-0.5cm},
			]
			\addplot coordinates {(What, 1698) (When, 2478) (How, 46) (How many, 1822) (Where, 1721) (Who, 1034) (Which, 1490)};
		\end{axis}
	\end{tikzpicture}
	
	\caption{Distribution of the questions by type}
	\label{fig:types}
\end{figure}
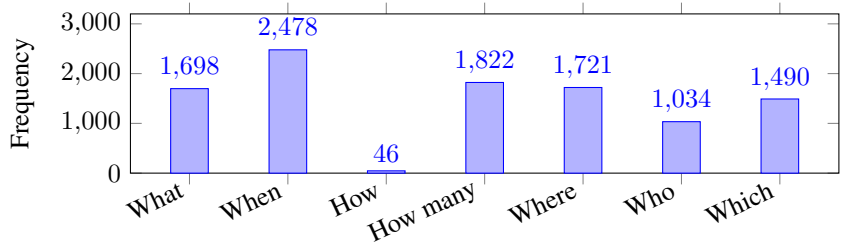

One of the key properties of an MCQ is its type. The type of question plays a crucial role in assessing different learning objectives and cognitive skills. Here, we analyze the distribution of questions in our dataset based on their information retrieval focus, using a rule-based approach that checks question words used in the question. The distribution of question types is shown in Figure \ref{fig:types}. The questions are categorized based on their type as follows:

\textbf{What:} This category covers questions that test factual information, asking about the nature or definition of something. 

\textbf{When:} These questions aim to assess knowledge of chronology or time-related information.

\textbf{How:} This category focuses on questions that explore the manner, quality, or degree of something.

\textbf{How Many:} This type targets questions about quantity or numerical information.

\textbf{Where:} These questions test knowledge about location.

\textbf{Who:} This category focuses on identifying people or entities associated with specific events or concepts.

\textbf{Which:} This type involves questions that require choosing the most appropriate option from a set of possibilities.

\subsection{Categorizing Based on Content}
\begin{figure*}[h]
	\centering
	\begin{tikzpicture}
		\begin{axis}[
			ybar,
			symbolic x coords={history, technology, health and medicine, economy and commerce, politics, geography, art and culture, science, sport, society, religion, others},
			ymin=0, ymax=2000,
			xtick=data,
			nodes near coords,
			bar width=0.5cm,
			width=13cm, height=4cm,
			xticklabel style={rotate=30, anchor=east},
			ylabel={Frequency},
			xlabel style={yshift=-0.5cm},
			]
			\addplot coordinates {(history, 1552) (technology, 305) (health and medicine, 316) (economy and commerce, 353) (politics, 469) (geography, 1276) (art and culture, 1054) (science, 1467) (sport, 1073) (society, 310) (religion, 1155) (others, 959)};
		\end{axis}
	\end{tikzpicture}
	\caption{Distribution of the questions by content}
	\label{fig:content-chart}
\end{figure*}
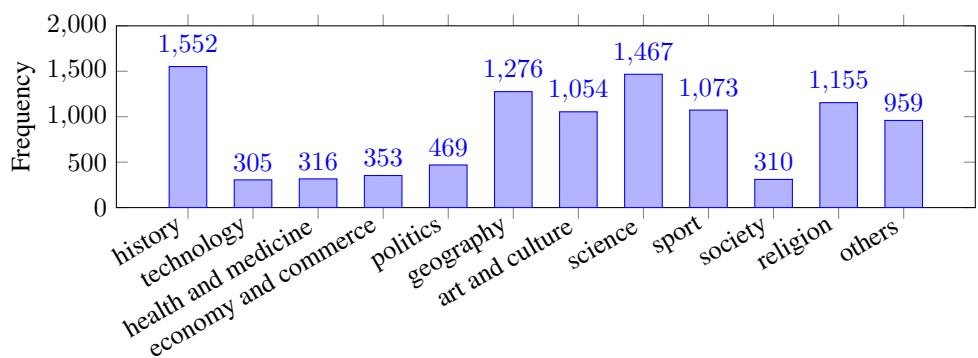

Another important characteristic of an MCQ lies in its content. Analyzing the content of questions helps to identify which categories the proposed model and multiple-choice question-answering models excel in, thereby enabling a targeted approach to enhance performance across other categories. We analyzed the distribution of questions based on their content, using GPT-4o API. The prompt used to categorize questions based on their content is presented in Figure \ref{fig:categorization-promt}. The distribution of different content categories is shown in Figure \ref{fig:content-chart}. The questions are categorized based on their content as follows:

\begin{figure*}[h]
	\centering
	\includegraphics[width=11cm, height=5.9cm]{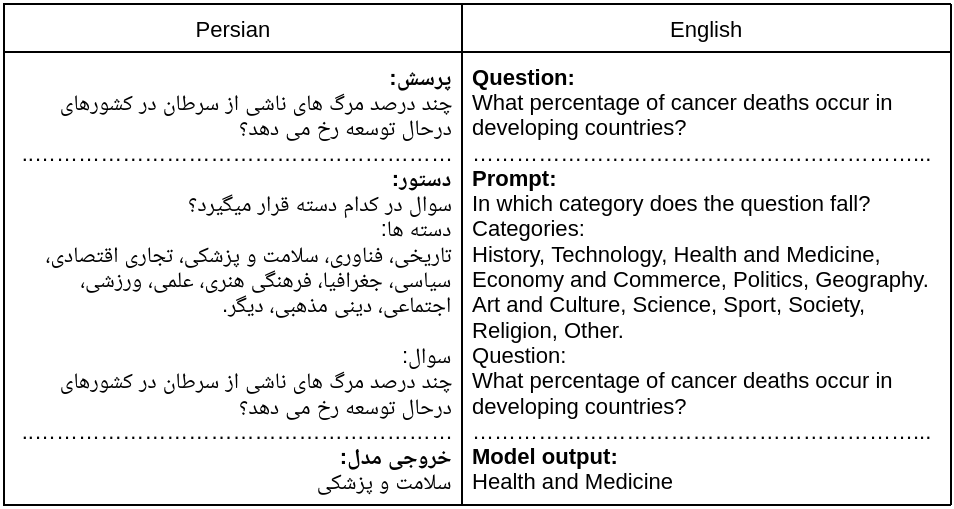}
	\caption{Prompt example to categorize questions based on their content. The left part is original Persian data and the right part presents its corresponding English translation.}
	\label{fig:categorization-promt}
\end{figure*}

\textbf{History:} This topic encompasses inquiries pertaining to historical occurrences and figures.

\textbf{Technology:} Questions within this domain are designed to gauge understanding of technological advancements, encompassing devices like smartphones, laptops, and computers, alongside their manufacturers examples include Apple and Samsung.

\textbf{Health and Medicine:} This category focuses on health and medical-related queries and delves into subjects such as diseases, human anatomy, treatments, and more.

\textbf{Economy and Commerce:} This category is designed to evaluate knowledge regarding economic activities, including production methods and commerce.

\textbf{Politics:} This topic contains questions focused on politics, including information about politicians, ideologies, governmental structures, and related concepts.

\textbf{Geography:} This category is dedicated to inquiries about various locations, ranging from natural landmarks to cityscapes and country positions.

\textbf{Art and Culture:} This topic features questions linked to artistic expressions, including paintings, movies, music, crafts, and more. 

\textbf{Science:} This category comprises questions centered around scientific disciplines, including chemistry, physics, and mathematics.

\textbf{Sport:} This category includes information about sports, highlighting notable individuals and events in this field, such as football and basketball.

\textbf{Society:} This category delves into societal aspects, covering topics like demographics, law enforcement, and community dynamics.

\textbf{Religion:} This topic addresses questions concerning religious doctrines and practices across faiths, including Zoroastrianism, Islam, Christianity, Judaism, Buddhism, etc.

\textbf{Others:} This final category encompasses questions with subjects not fitting into the aforementioned categories, such as biography, translation, etc.

\section{Evaluation}
\label{sec:eval}
To assess the quality of the questions generated with the proposed pipeline, we employed both automated and human evaluation methods. For automated evaluation, we leveraged \texttt{Dorna2-Llama3.1-8B-Instruct} (Dorna2) \citep{partai-dorna2-llama31-8b-instruct-2025}, \texttt{Maral-7B-alpha-1} (Maral-7B) \citep{maralgpt-maral7b-alpha1}, \texttt{Meta-Llama-3.1-8B-Instruct} (Llama-8B) \citep{unsloth-llama3-8b-2025}, \texttt{Mistral-7B-Instruct-v0.3} (Mistral-7B) \citep{mistral7b2024}, \texttt{Qwen2.5-7B-Instruct} (Qwen-7B) and \texttt{Qwen2.5-14B-Instruct} (Qwen-14B) \citep{qwen2.5}, measuring their accuracy and confidence in selecting the correct answers and the probability they assign to each choice using Lm-Evaluation-Harness \citep{gao-2024-lm-eval-harness}. This helps gauge the ambiguity and effectiveness of the generated wrong choices. Additionally, we conducted a human evaluation on a sample of 200 questions, assessing two key metrics: (1) Validness—whether the questions and their options are logically valid, and (2) Distractiveness—the ability of wrong choices to plausibly mislead examinees. The human evaluation was conducted by two undergraduate students who are native Persian speakers.

To evaluate the performance of LLMs based on the probabilities assigned to each choice, we employed four metrics. Hard Accuracy (Equation \ref{eq:hard-accuracy}) measures the proportion of correctly answered questions relative to the total number of questions. Soft Accuracy (Equation \ref{eq:soft-accuracy}) represents the average probability assigned by the LLM to the correct choice across all questions. The Confidence of an LLM for each question is defined in Equation \ref{eq:confidence}, while the Normalized Entropy (Equation \ref{eq:normalized-entropy}) quantifies the uncertainty of the model’s predictions by dividing the entropy of the choice probabilities by the logarithm of the number of choices (in this case, 4). Mean Confidence (Equation \ref{eq:mean-confidence}) denotes the average confidence of the LLM across all questions in the dataset. Finally, Correlation (Equation \ref{eq:correlation}) is computed between the Confidence and Soft Accuracy values for each question answered by the LLM. All numerical values in the tables are scaled by 100 for better readability.

\begin{equation}
	Hard\,Accuracy = \frac{Correctly\:Answered}{Total}
	\label{eq:hard-accuracy}
\end{equation}

\begin{equation}
	Soft\,Accuracy = \frac{\sum_{i=1}^{n}Correct\,Choice\,Prob}{Total}
	\label{eq:soft-accuracy}
\end{equation}

\begin{equation}
	Normalized\,Entropy = -\frac{1}{\log c} \sum_{i=1}^{c} p_i \log p_i
	\label{eq:normalized-entropy}
\end{equation}

\begin{equation}
	Confidence = 1 - Normalized\,Entropy
	\label{eq:confidence}
\end{equation}

\begin{equation}
	Mean\,Confidence = \frac{\sum_{i=1}^{n} Confidence(Q_i)}{Total}
	\label{eq:mean-confidence}
\end{equation}

\begin{equation}
	Correlation = \frac{\sum_{i=1}^{n} (x_i - \bar{x})(y_i - \bar{y})}
	{\sqrt{\sum_{i=1}^{n} (x_i - \bar{x})^2} \sqrt{\sum_{i=1}^{n} (y_i - \bar{y})^2}}
	\label{eq:correlation}
\end{equation}

Table \ref{tab:llm_performance} presents the quantitative comparison of six LLMs under both 16-bit and 8-bit quantization settings. The difference between 16-bit and 8-bit quantization is minimal for all models, indicating that quantization has a negligible impact on both accuracy and confidence. Among the evaluated models, Llama-8B and Qwen-14B achieved the highest hard and soft accuracies, followed closely by Dorna2. In contrast, Maral-7B and Mistral-7B demonstrated lower accuracies. The confidence results further revealed significant differences among models. Maral-7B exhibited overconfidence despite lower accuracy, whereas Qwen-14B models maintained more conservative confidence levels. The correlation metric confirmed that Llama-8B and Dorna2 provided the most reliable confidence estimates, reflecting a stronger alignment between certainty and correctness.

\begin{table*}[h]
	\centering
	\caption{Performance comparison of LLMs in 16-bit and 8-bit quantized formats.}
	
	\begin{tabular}{lcccccccc}
		\hline
		\textbf{Model} & \textbf{Quant.} & \textbf{Hard Acc.} & \textbf{Soft Acc.} & \textbf{Mean Conf.} & \textbf{Correlation} \\
		\hline
		Dorna2-Llama3.1-8B-Instruct & 16-bit & 43.9 & 39.2 & 71.8 & 69.3 \\
		& 8-bit  & 43.7 & 39.1 & 71.7 & 68.9 \\
		\hline
		Maral-7B-alpha-1            & 16-bit & 31.1 & 27.2 & 95.9 & 56.3 \\
		& 8-bit  & 30.4 & 27.0 & 96.0 & 54.8 \\
		\hline
		Meta-Llama-3.1-8B-Instruct  & 16-bit & 46.0 & 40.1 & 74.4 & 74.6 \\
		& 8-bit  & 45.3 & 40.1 & 73.8 & 73.5 \\
		\hline
		Mistral-7B-Instruct-v0.3    & 16-bit & 29.5 & 28.4 & 72.2 & 22.4 \\
		& 8-bit  & 29.6 & 28.3 & 73.1 & 23.4 \\
		\hline
		Qwen2.5-14B-Instruct        & 16-bit & 49.2 & 48.8 & 25.3 & 40.7 \\
		& 8-bit  & 48.1 & 47.9 & 27.2 & 40.8 \\
		\hline
		Qwen2.5-7B-Instruct         & 16-bit & 43.2 & 42.5 & 30.5 & 34.8 \\
		& 8-bit  & 42.6 & 42.2 & 31.3 & 34.9 \\
		\hline
	\end{tabular}
	\label{tab:llm_performance}
\end{table*}

Table \ref{tab:llm_accuracy_types} reports the Soft Accuracy of each LLM across different question types in the generated dataset. The results show that model performance varies notably depending on the question type. Qwen-14B achieved the highest accuracy across all types, particularly excelling in Where and How questions (62.6\% and 71.3\% in 16-bit, respectively). Qwen-7B, Llama-8B and Dorna2 followed closely, maintaining strong performance in Where, How, and What types, while Maral-7B and Mistral-7B consistently produced lower scores across all types. The How and Where questions appeared generally easier for most models, whereas When and How Many were more challenging. The minimal differences between 16-bit and 8-bit quantization once again confirm that quantization has negligible influence on model accuracy across different question types.

\begin{table*}[h]
	\centering
	\caption{Soft Accuracy of each LLM across different question types}
	\resizebox{\textwidth}{!}{%
		\begin{tabular}{lcccccccc}
			\hline
			\textbf{Model} & \textbf{Quant.} & \textbf{When} & \textbf{How Many} & \textbf{Where} & \textbf{What} & \textbf{Which} & \textbf{Who} & \textbf{How} \\
			\hline
			Dorna2-Llama3.1-8B-Instruct       & 16-bit & 29.5 & 27.4 & 51.0 & 47.4 & 43.5 & 43.1 & 48.9 \\
			& 8-bit  & 29.5 & 27.5 & 51.1 & 47.1 & 43.5 & 42.9 & 49.7 \\
			\hline
			Maral-7B-alpha-1                  & 16-bit & 25.8 & 25.2 & 29.4 & 29.4 & 27.4 & 26.4 & 30.4 \\
			& 8-bit  & 25.7 & 25.1 & 29.1 & 29.2 & 27.1 & 26.3 & 30.7 \\
			\hline
			Meta-Llama-3.1-8B-Instruct        & 16-bit & 30.0 & 27.0 & 52.5 & 49.4 & 44.7 & 44.8 & 52.0 \\
			& 8-bit  & 30.0 & 27.1 & 52.4 & 49.2 & 44.6 & 45.0 & 52.6 \\
			\hline
			Mistral-7B-Instruct-v0.3          & 16-bit & 24.6 & 25.6 & 32.2 & 33.4 & 28.9 & 26.3 & 41.2 \\
			& 8-bit  & 24.8 & 25.6 & 32.0 & 33.2 & 28.8 & 26.3 & 40.2 \\
			\hline
			Qwen2.5-14B-Instruct              & 16-bit & 35.0 & 34.0 & 62.6 & 59.2 & 54.7 & 58.3 & 71.3 \\
			& 8-bit  & 34.1 & 34.5 & 61.0 & 57.7 & 54.4 & 56.8 & 63.9 \\
			\hline
			Qwen2.5-7B-Instruct               & 16-bit & 30.5 & 33.1 & 53.7 & 51.5 & 47.2 & 46.9 & 61.9 \\
			& 8-bit  & 30.4 & 32.8 & 53.6 & 50.2 & 47.0 & 46.9 & 59.2 \\
			\hline
		\end{tabular}
	}
	\label{tab:llm_accuracy_types}
\end{table*}

Table \ref{tab:llm_accuracy_content} illustrates the Soft Accuracy of LLMs across different content categories. Qwen2.5-14B consistently outperforms other models, particularly on Society (55.8\%), Geography (53.5\%), and Politics (52.3\%) questions, while Qwen2.5-7B generally ranks second, with its highest scores in Technology (48.3\%) and Society (46.8\%). Dorna2 and Llama-8B achieve moderate accuracy, peaking in Geography (46.2\% and 47.4\%, respectively) and Politics (42.1\% and 43.0\%, respectively). In contrast, Maral-7B and Mistral-7B exhibit consistently lower performance across all categories, rarely exceeding 30\% in any content category. Across all models, Geography, Society, and Politics questions tend to yield higher accuracy, whereas History and Sport are more challenging, resulting in lower Soft Accuracy.

\begin{table*}[h]
	\centering
	\scriptsize
	\caption{Soft Accuracy of each LLM across different question content categories}
	\resizebox{\textwidth}{!}{%
		\begin{tabular}{llcccccccccccc}
			\hline
			\textbf{Model} & \textbf{Quant.} & \textbf{History} & \textbf{Tech.} & \textbf{Health} & \textbf{Econ.} & \textbf{Politics} & \textbf{Geo.} & \textbf{Art} & \textbf{Science} & \textbf{Sport} & \textbf{Society} & \textbf{Religion} & \textbf{Other} \\
			\hline
			Dorna2-Llama3.1-8B-Instruct & 16-bit & 35.6 & 42.9 & 39.6 & 38.1 & 42.1 & 46.2 & 38.1 & 38.5 & 34.3 & 41.6 & 39.8 & 39.3 \\
			& 8-bit  & 35.6 & 42.5 & 39.2 & 38.4 & 42.1 & 46.2 & 38.0 & 38.3 & 34.3 & 41.8 & 39.9 & 39.3 \\
			\hline
			Maral-7B-alpha-1 & 16-bit & 26.0 & 28.1 & 26.3 & 27.9 & 26.8 & 28.3 & 28.0 & 27.3 & 26.3 & 28.7 & 27.0 & 27.3 \\
			& 8-bit  & 25.8 & 27.7 & 26.3 & 27.9 & 26.6 & 28.1 & 27.8 & 27.1 & 26.2 & 28.5 & 26.9 & 27.2 \\
			\hline
			Meta-Llama-3.1-8B-Instruct & 16-bit & 36.9 & 42.3 & 41.5 & 38.5 & 43.0 & 47.4 & 39.2 & 38.7 & 35.3 & 43.9 & 40.7 & 40.6 \\
			& 8-bit  & 36.8 & 42.3 & 41.0 & 38.8 & 43.4 & 47.4 & 39.3 & 38.7 & 35.2 & 43.9 & 40.5 & 40.4 \\
			\hline
			Mistral-7B-Instruct-v0.3 & 16-bit & 25.2 & 30.3 & 27.5 & 29.0 & 27.9 & 30.9 & 29.6 & 28.9 & 27.3 & 30.8 & 27.4 & 29.2 \\
			& 8-bit  & 25.3 & 30.9 & 27.5 & 29.2 & 27.9 & 30.4 & 29.6 & 28.8 & 26.9 & 31.0 & 27.8 & 29.2 \\
			\hline
			Qwen2.5-14B-Instruct & 16-bit & 44.0 & 51.7 & 51.5 & 48.1 & 52.3 & 53.5 & 47.4 & 47.0 & 42.9 & 55.8 & 50.1 & 54.0 \\
			& 8-bit  & 43.0 & 51.1 & 50.0 & 47.2 & 51.5 & 52.6 & 47.4 & 46.4 & 42.6 & 54.2 & 49.2 & 51.8 \\
			\hline
			Qwen2.5-7B-Instruct & 16-bit & 37.4 & 48.3 & 41.1 & 41.5 & 45.8 & 46.7 & 40.9 & 45.0 & 37.8 & 46.8 & 41.7 & 45.5 \\
			& 8-bit  & 37.3 & 48.4 & 39.8 & 41.2 & 45.1 & 46.7 & 40.2 & 44.3 & 37.6 & 46.6 & 41.5 & 45.2 \\
			\hline
		\end{tabular}
	}
	\label{tab:llm_accuracy_content}
\end{table*}

The human evaluation results indicate that 97.5\% of the generated questions are valid and have valid choices, and 94.5\% of the generated wrong choices are effectively distractive, underscoring the capability of the proposed pipeline to produce both accurate and challenging questions.

The analysis of the performance of LLMs, together with the human evaluation, demonstrates that the proposed pipeline for generating MCQs was effective and produced a high-quality dataset.

\section{Conclusion}
\label{sec:conclusion}

In this study, we presented a framework for generating MCQs in Persian. The framework employs an mT5-based model for question generation and leverages GloVe, Word2Vec, and BERT-based fill-mask models to produce plausible wrong choices. To improve quality, the generated options are filtered using part-of-speech and named-entity tags, as well as their written forms. A knowledge graph and BERT embeddings are then used to rank and select the most appropriate generated choices. The resulting dataset was evaluated using six LLMs and a human evaluation of 200 samples. Both assessments confirmed that the generated questions and options are valid and effective. These results demonstrate the potential of our approach to enhance automatic MCQ generation for the Persian language and provide a valuable resource for future research in educational technology and language processing.

\section{Limitations}
Although the proposed framework effectively generates MCQs with plausible choices, certain limitations should be acknowledged. First, the quality of the generated questions and choices depends on the performance of the underlying language and embedding models—such as mT5, GloVe, and Word2Vec—which may not fully capture fine-grained semantic nuances, especially in domain-specific contexts. Second, due to limited computational resources, the dataset was evaluated using only six large language models, most of which contain 7–8 billion parameters, with only one model having 14 billion parameters. This constraint may limit the generalizability of the results to larger or more diverse models.

\section{Acknowledgements}
We would like to express our sincere gratitude to Mr. Farhan Farsi for his valuable guidance throughout this project.

\bibliographystyle{unsrtnat}
\bibliography{references}  






\end{document}